\def\year{2017}\relax
\documentclass[letterpaper]{article}
\usepackage{aaai17}
\usepackage{times}
\usepackage{helvet}
\usepackage{graphicx}
\usepackage{amsmath}
\usepackage{amssymb}
\usepackage{courier}
\begin{document}
%
\title{Fine-Grained Car Detection for Visual Census Estimation}
\author{Timnit Gebru \and Jonathan Krause \and Yilun Wang \and Duyun Chen \and Jia Deng \and Li Fei-Fei\\
Department of Computer Science, Stanford University\\
{\texttt {\{tgebru, jkrause, yilunw, duchen, feifeili\}@cs.stanford.edu}}\\
Department of Computer Science, University of Michigan\\
\texttt {jiadeng@umich.edu}
}

\maketitle

\begin{abstract}
Targeted socio-economic policies require an accurate understanding of a country's demographic makeup. To that end, the United States spends more than 1 billion dollars a year gathering census data such as race, gender, education, occupation and unemployment rates. Compared to the traditional method of collecting surveys across many years which is costly and labor intensive, data-driven, machine learning-driven approaches are cheaper and faster\textemdash with the potential ability to detect trends in close to real time. In this work, we leverage the ubiquity of Google Street View images and develop a computer vision pipeline to predict income, per capita carbon emission, crime rates and other city attributes from a single source of publicly available visual data. We first detect cars in 50 million images across 200 of the largest US cities and train a model to predict demographic attributes using the detected cars. To facilitate our work, we have collected the largest and most challenging fine-grained dataset reported to date consisting of over 2600 classes of cars comprised of images from Google Street View and other web sources, classified by car experts to account for even the most subtle of visual differences. We use this data to construct the largest scale fine-grained detection system reported to date. Our prediction results correlate well with ground truth income data (r=0.82), Massachusetts department of vehicle registration, and sources investigating crime rates, income segregation, per capita carbon emission, and other market research. Finally, we learn interesting relationships between cars and neighbourhoods allowing us to perform the first large scale sociological analysis of cities using computer vision techniques.

\end{abstract}

\begin{figure*}[t]
\begin{center}
   \includegraphics[width=1\linewidth]{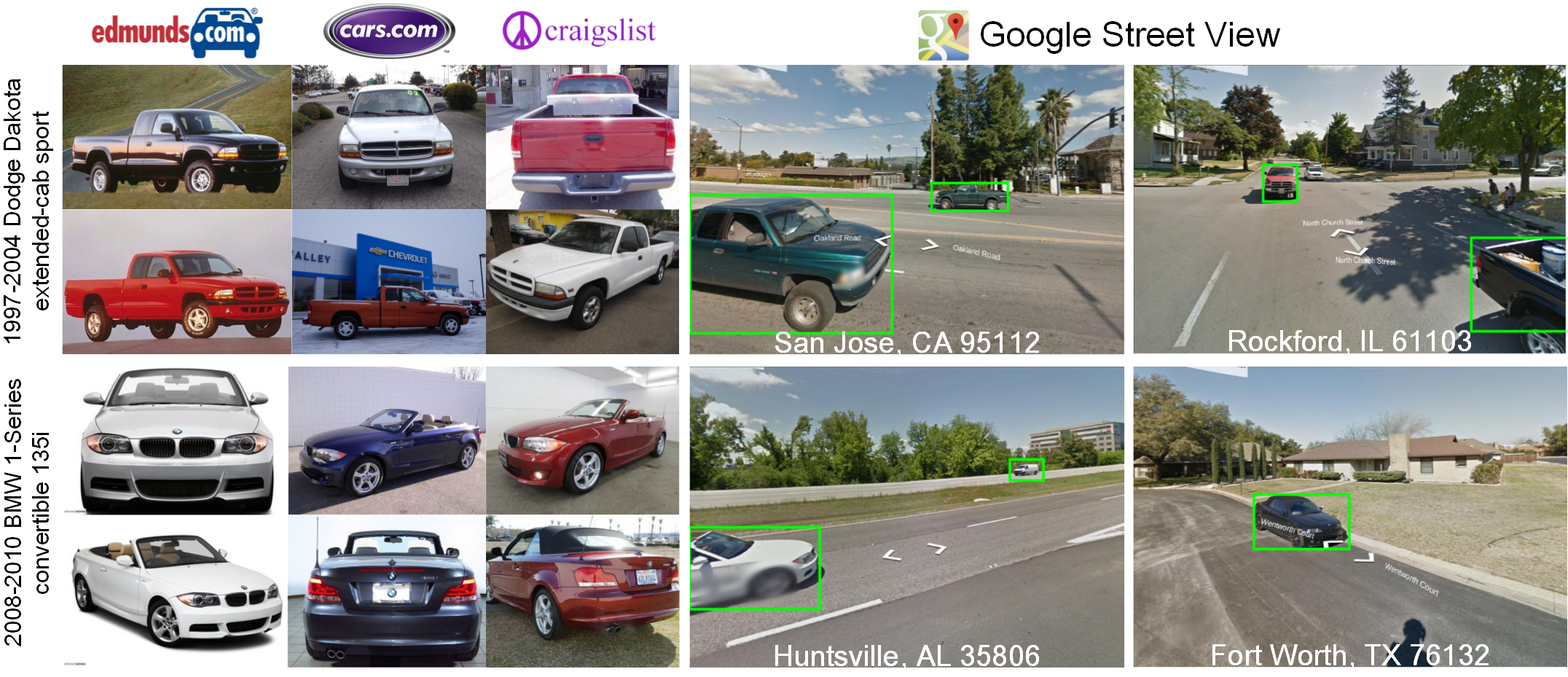}
\end{center}
   \caption{ Examples of cars from our fine-grained car dataset. Left: examples of cars from edmunds.com, cars.com and craigslist.com. Right: examples of cars from streetview images. Green bounding boxes indicate the ground truth location of cars in Street View images.}
\label{fig:dataset1}
\end{figure*}
\section{Introduction}

Many government and non-government projects are dedicated to studying cities and their inhabitants. For example, the American Community Survey (ACS) collects data related to the demographic makeup of the US, the Environmental Protection Agency gathers data pertaining to city pollution, and private organizations such as car dealerships gather information regarding the relationship between cars and demographics. Traditionally, the most prevalent method for obtaining such personal, demographic and environmental information is through costly surveys such as the US census, American community survey (ACS) and other projects conducted by disparate government entities. However, the emergence of large, diverse sets of data generated by people has enabled computer scientists and computational sociologists to gain interesting insights by analyzing massive user texts and social networks~\cite{jure,nlp_people}. For instance Michel et al analyzed over one million books and presented results related to the evolution of the English language as well as various cultural phenomena~\cite{ngrams}.  And  Blumenstock et al used mobile phone data to predict poverty rates in Rwanda ~\cite{mobile}.

In contrast to textual data, the use of images for computational social science has been largely unexplored. A snapshot of a street or neighborhood tells a detailed story of its socioeconomic make up. However, while a few pioneering works have applied visual scene analysis techniques to infer characteristics of neighborhoods and cities, they have been limited in scope and scale. Jean et al used Satellite image features to predict poverty rates in 5 African countries~\cite{neal}. Similarly, works from Zhou et al, Ordonez et al, Naik et al and Arietta et al use global image features from Google Street View images to learn various neighborhood characteristics~\cite{antonio,tamara,mit_cvpr,alyosha}.

Global image features, however, contain limited information about individuals, neighborhoods and cities. In contrast, affluence (or lack thereof), culture, and even crime can be inferred by observing houses, people, clothing styles and types of cars on the street. Ninety five percent of American households own automobiles~\cite{car_stats}, and as shown by prior work ~\cite{car_personality} cars are a reflection of their owners' characteristics providing significant personal information. For instance, a large number of Teslas is a strong indicator of a wealthy neighborhood.

Guided by this intuition we develop a novel computer vision pipeline to predict demographic variables from Google Street View images. We recognize cars in 50 million images from 200 of the largest American cities and use the detected cars to predict  income, segregation levels, per capita carbon emission and crime rates. In addition to accurately predicting ground truth data (e.g. Pearson’s r=0.82 for income), we discover interesting relationships between cars and demographics allowing us to perform a sociological study of our 200 cities. For instance, we find that wealthy people drive foreign made cars and that a large quantity of vans is correlated with high crime.

While specific vehicle attributes can be associated with certain types of people, classifying different types of cars is a difficult computer vision task called fine-grained object recognition. In fine-grained recognition we seek to distinguish between similar objects such as different types of cars, dog breeds or clothing styles\textemdash a task that can even be challenging for humans. In this work, we train a fast, large scale fine-grained detection system able to detect cars in 50 million Google Street View images in less than 2 weeks. To train our model we gathered a fine-grained car dataset of unprecedented scale. It has 2657 car classes consisting of nearly all car types produced in the world after 1990: with a total of 700,000 images from websites such as edmunds.com, cars.com, craigslist.com and Google Street View (Fig.~\ref{fig:dataset1}). We make our dataset publicly available and anticipate its use by computer vision researchers focused on fine-grained recognition. 

Finally, we use the detected cars to train a model predicting socioeconomic attributes such as income and segregation levels. We show that from a single source of publicly available visual data, Google Street View images, we are able to predict diverse sets of important societal information typically gathered by different entities. We not only show good correlation with socioeconomic census data but can also predict information absent from the census such as city pollution levels, crime rates, income segregation levels, vehicle registration and other data related to cities.

\section{A Large Scale Fine-Grained Car Dataset}

\begin{figure*} [t]
\begin{center}
 \includegraphics[width=1\linewidth]{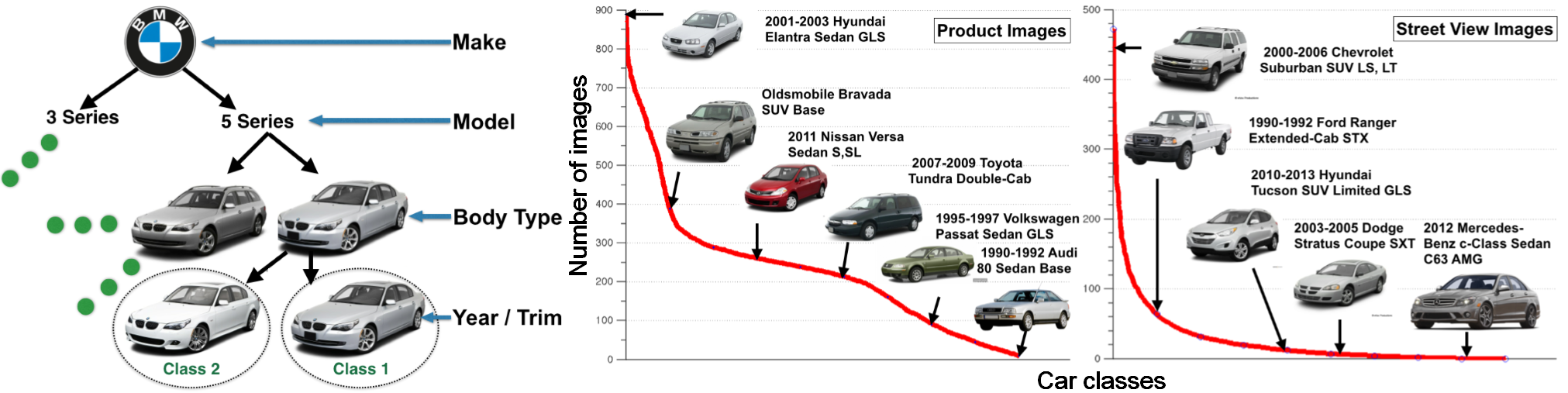}
\end{center}
\caption {Left: A hierarchy of car classes in our dataset. Classes become more difficult to distinguish lower in the hierarchy, with
differences extremely subtle at the year and trim level. Center: The number of images per class obtained from edmunds.com, cars.com,
and craigslist.com vs. Right: The number of images per class from Street View.}
\label{fig:img_dist}
\end{figure*}

\begin{table}
\begin{center}
\begin{tabular}{|l|c|c|c|}
\hline
\textbf{Attribute} & \textbf{Training} & \textbf{Validation} & \textbf{Test} \\
\hline\hline
Street View Images & 199,666 & 39,933 &159,732\\
Product Shot Images & 313,099 & - &-\\
\hline\hline
Total Images &512,765&39,933&159,732\\
\hline\hline
Street View BBoxes & 34,712 & 6,915 &27,865\\
Product Shot BBoxes & 313,099 & - &-\\
\hline\hline
Total BBoxes & 347,811 & 6,915 &27,865\\
\hline
\end{tabular}
\end{center}
\caption{Dataset statistics for our training, validation, and test splits. ``BBox'' is shorthand for Bounding Box. Product shot bounding boxes and images are from craigslist.com, cars.com and edmunds.com.}
\label{table:data-stat}
\end{table}

In order to analyze society via the cars visible in it, one must first create a dataset of all possible cars we would like to study. However, this poses a challenge. What are all of the cars in the world, and how can we possibly collect data for each one of them? We cannot use existing datasets such as~\cite{imagenet} as they do not have an exhaustive set of annotated car images.
To tackle this problem, we leverage existing knowledgebases of cars, downloading images and data for roughly 18,000 different types of cars from the website edmunds.com. We then grouped these car types into sets of indistinguishable classes and collected more images for each class from craigslist.com and cars.com. Both websites contain ground truth car images from owners looking to sell their cars.  Using Amazon Mechanical Turk (AMT), we collected one bounding box for each image yielding 2,657 visual groups of cars for our analysis.  A bounding box is a set of pixel coordinates describing the location of the image containing the object of interest\textemdash in our case a car. 

However, these images alone are not enough for an accurate study of society via cars visible on the street. To perform our social analysis, we need to recognize cars appearing in Google Street View images. As shown in (Fig.~\ref{fig:dataset1}), product shot images typically focus on a single car, centrally placed, and from diverse viewpoints. Google Street View images, on the otherhand, may contain multiple cars per image, each of which can be occluded and low resolution. This presents a challenge for an automobile recognition system. Thus, as part of our fine-grained car dataset we also collected bounding boxes for nearly 400,000 cars in Street View images. These were annotated by expert human labelers with one of the 2,657 fine-grained classes, providing us with data to train detectors and classifiers effective on real-world images. In Tab. 1 we present aggregate statistics of our fine-grained car dataset, which has a total of 712,430 images and 382,591 bounding boxes. 

What does this data look like? We analyze our car classes in Fig.~\ref{fig:img_dist}. At the most fine-grained level of car year and trim, differences between classes become extremely subtle. A plot of the class distribution further shows that product images have coverage over a wide variety of classes while the Street View distribution is much more skewed. Finally, to perform our visual census, we collected 50 million Google Street View images, sampling 8 million GPS points in 200 large US cities every 25m. Images from 6 camera rotations were captured per GPS point. 

\section{Fine-Grained Detection and Classification}
\label{sec:detection}
We train a model to distinguish cars in Google Street View images in 2 steps. First, we localize parts of the image having a high likelihood of containing a car; this is the generic car detection step. Then we take the regions with the highest likelihood and classify the types of cars they contain using a custom convolutional neural network (CNN). 

\subsection{Car Detection}
Our goal here is to train a model that can efficiently and accurately detect cars in 50 million images with reasonable accuracy. To this end, we are willing to sacrifice a couple of percent in accuracy to significantly speed up train and test times. Thus, although the current state-of-the-art for object detection is faster R-CNN~\cite{frcnn} we choose the simplicity and efficiency of deformable part models(DPM)~\cite{dpm} for our task of large scale car detection. 
After extensive cross validation, we decided upon a single component DPM with 8 parts, achieving an average precision (AP) of 64.2\% at 5 seconds per Street View image. Given an input image, our trained DPM outputs bounding boxes of varying sizes in the image with associated scores reflecting the likelihood of containing a car. With this architecture, we detected cars on our entire dataset in less than two weeks with 200 2.1 GHz CPU cores.

To further improve detection performance, we leveraged knowledge of car distributions in Street View images. This is done by introducing a prior on the location and size of predicted bounding boxes. Concretely, we model the distribution of Street View cars in images as a histogram over three variables: the $x$ coordinate of the center of the bounding box, the $y$ coordinate of the center, and $log(area)$ where $area$ is the area of the bounding box. We divide each variable into 20 bins resulting in a total of 8,000 bins in the histogram, and estimate the probability of each bin using bounding box statistics in the Street View training data. We add a pseudo count of 1 in each bin for regularization. With $P(x,y,\log{(area)})$ denoting this histogram, we augment each DPM detection score by:

\begin{equation}
score^{DPM+LOC} = score^{DPM} + \alpha \log{(P(x,y,\log{(area)}))}
\end{equation}
where $\alpha$ is a learned weight on the location prior. This improves detection AP by 1.92 at a negligible time cost. During analysis, DPM scores are converted into estimated probabilities via isotonic regression~\cite{isotonic}, learned on the validation set.
The final output of the detection step provides us with a set of bounding boxes with varying sizes and locations in an image, and the probability that each of these boxes contains a car.

\subsection{Car Classification}
To classify the detected cars into one of 2,657 fine-grained classes, we use a convolutional neural network with an architecture following ~\cite{alexnet}. We choose this architecture for its efficiency instead of state-of-the-art fine-grained classification systems like~\cite{jon1,jon2}. Discriminative learning methods including CNNs work best when trained on data from a similar distribution as the test set (in our case, Street View images)~\cite{domain}. However, the prohibitive cost of labeling many Street View images for each of our 2,657 categories prevents us from having enough training data from this source. Instead, we train our CNN on a combination of Street View and the more plentiful and inexpensively labeled product shot images. We made three modifications to the traditional CNN training procedure to improve our classifier performance.

First, we seek to prevent our classifier from overfitting on product shot images since they are a much larger fraction of our training set. Inspired by domain adaptation works, we approximate the WEIGHTED method of Daume Ì ~\cite{frustrating} by duplicating each Street View example 10 times during training. This roughly equalizes the number of product shot and Street View training images. Next, we apply transformations to product shot images to make them similar to those from Street View. This decreases the distance between training and testing data distributions, improving classifier performance. Cars in product shot images occupy a much larger number of pixels in the image (Fig.~\ref{fig:dataset1}) than those in Street View. To compensate, we first measured the distribution of bounding box resolutions in Street View training images. At training time, we dynamically downsize each product shot image according to this distribution and  rescale it to fit the input dimensions of the CNN. Resolutions are parameterized by the geometric mean of the bounding box width and height, and the probability distribution is given as a histogram over 35 different such resolutions. 

A further challenge in classifying Street View images is that the input to our CNN consists of noisy bounding boxes output by the previous detection step. This stands in contrast to the default classifier input\textemdash ground truth bounding boxes that are tight around each car. To tackle this challenge, we use our validation data to measure the distribution of intersection over union (IOU) overlap between ground truth bounding boxes and those produced by our car detector. For each Street View bounding box input to the CNN, we randomly sample its source image according to this IOU distribution, simulating noisy detections during training.

At test time, each detected bounding box is input to the CNN, and we perform a single forward pass to get the softmax probabilities specifying the likelihood of containing a car belonging to each of the 2,657 categories. In practice, we only keep the top 20 predictions, since storing a full 2,657-dimensional floating point vector for each bounding box is prohibitively expensive. On average, these top 20 predictions account for 85.5\% of the softmax layer probability mass. We also note that, after extensive code optimization to make this classification step as fast as possible, we are primarily limited by the time spent reading images from disk, especially when using multiple GPUs simultaneously.

\begin{table}
\begin{center}
\begin{tabular}{|l|c|}
\hline
\textbf{Attribute} & \textbf{Accuracy} \\
\hline\hline
Make & 66.38\%\\
Model & 51.83\% \\
Submodel & 77.74\% \\
Price & 61.61\% \\
Domestic/Foreign & 87.71\%\\
Country & 84.21\%\\
\hline
\end{tabular}
\end{center}
\caption{Classification accuracy on the test set for various car attributes.}
\label{table:att-acc}
\end{table}

\begin{figure}[t]
\begin{center}
   \includegraphics[width=1\linewidth]{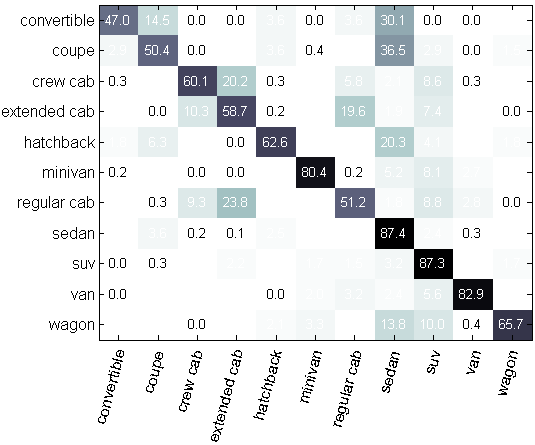}
\end{center}
   \caption{Confusion matrix for car body types. Each row represents
the ground truth body type and each column shows the body type predicted by our classifier.} 
\label{fig:confusion}
\end{figure}

\subsection{Analyzing Classification Performance}
Fine-grained classification across 2,657 classes is a challenging task even for expert humans. Our CNN achieves a remarkable classification accuracy of 31.27\%, on ground truth bounding boxes, and 33.27\% on true positive DPM detections. 
Since our model will not always be accurate at this level of granularity, we closely examine its errors. Some types of mistakes can undermine the accuracy of our visual census where as others do not matter. For example, misclassifying a 2001 Honda Accord LX as a 2001 Honda Accord DX does not affect our social analysis. But incorrectly classifying a 2012 BMW 3-Series as a 1996 Honda Accord could seriously impact the quality of our results. We list classification accuracy for various car attributes in Tab.~\ref{table:att-acc}. We can classify these car properties with much higher accuracies than the 2,657-way classification, indicating that errors significantly impacting the visual census are rare. We zoom in on the â€œBody Typeâ€ attribute in Fig.~\ref{fig:confusion} and observe that most errors are between highly similar body types; e.g. sedan and coupe, or extended and crew cab trucks.

 \section{Visual Census of Cities and Neighborhoods}
 \label{sec:census}
\begin{figure}[t]
\begin{center}
    \includegraphics[width=1.0\linewidth]{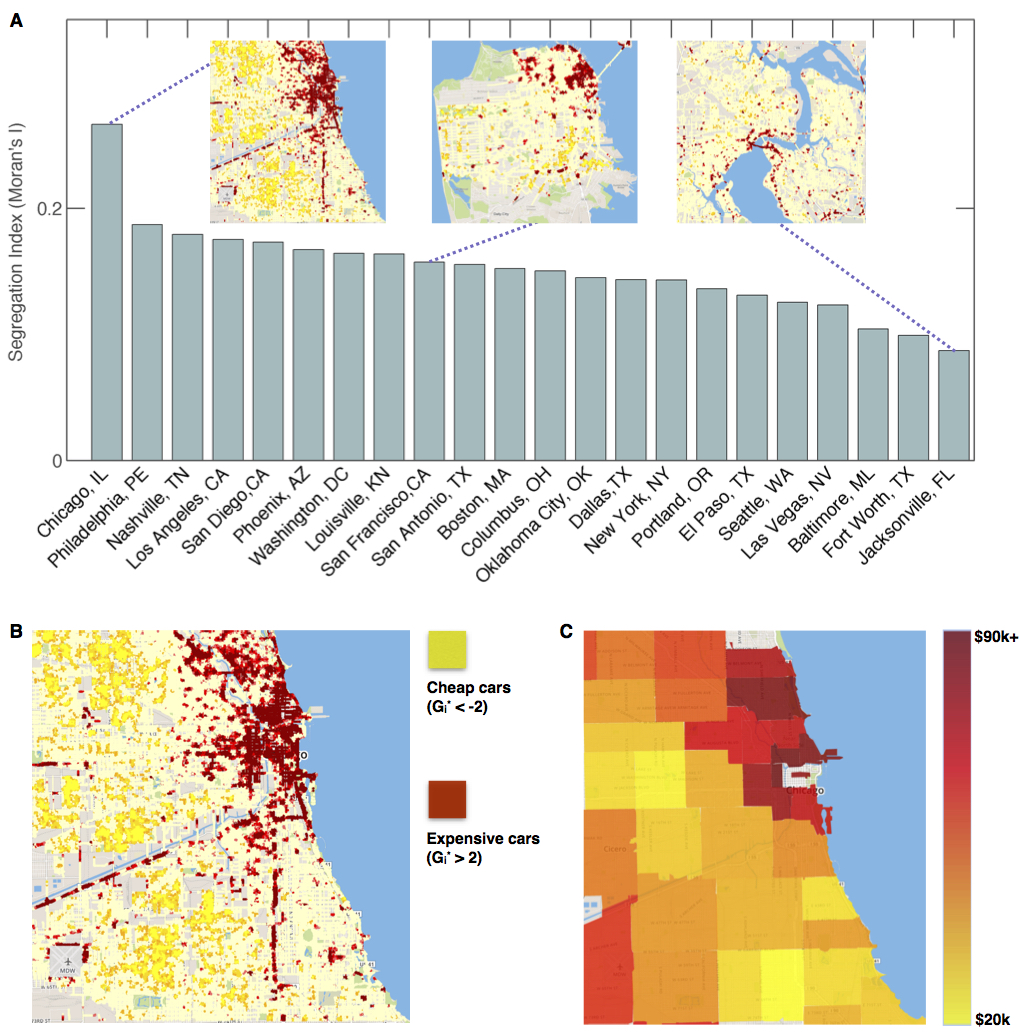}
\end{center}
   \caption {(A) Twenty two major American cities ranked by segregation of car price. Our segregation index is Moran's I statistic~\cite{moran}. Insets show maps of statistically significant clusters of cars with high prices (red), low prices (yellow) as well as no statistically significant clustering (white) for the cities of Chicago, IL, San Francisco, CA and Jacksonville, FL respectively using the Getis-Ord Gi\* statistic~\cite{getis}. Chicago has large clusters of expensive and cheap cars whereas Jacksonville shows almost no clustering of cars by price. (B) Expensive/Cheap clusters of cars in Chicago. (C) Zip code level median household income in Chicago. Large clusters of expensive cars are in wealthy neighborhoods whereas large clusters of cheap cars are in unwealthy neighborhoods. } 
\label{fig:moran-i}
\end{figure}

\begin{figure*} [t]
\begin{center}
\includegraphics[width=1.0\linewidth]{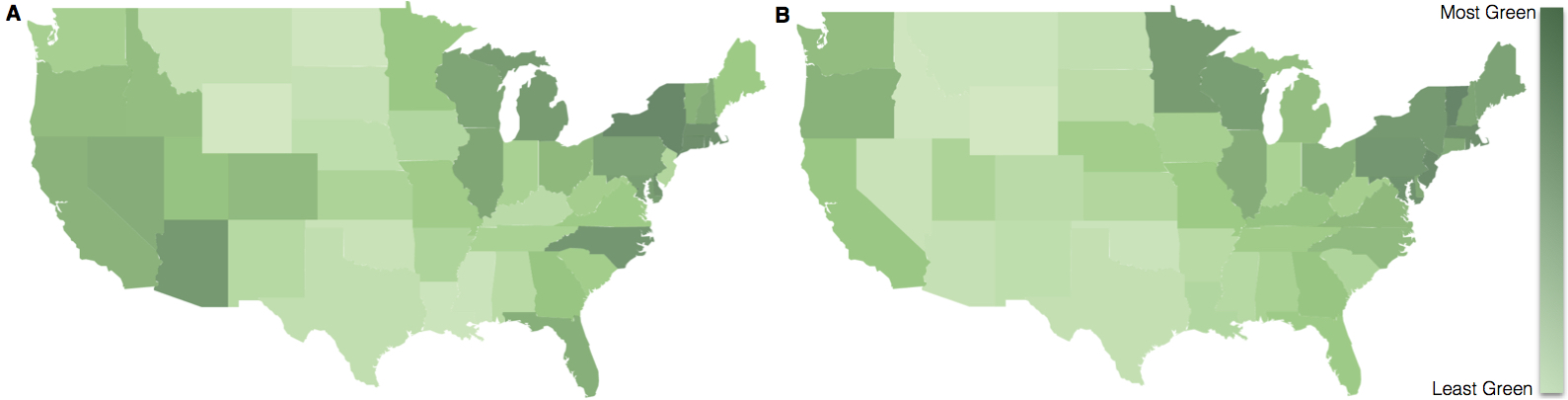}
\end{center}
\caption {(A) A map ranking each state's carbon footprint from the transportation sector in 2012 using data from~\cite{state_ranking}. (B) A map ranking each state's average miles per gallon (MPG) calculated from car attributes detected from Google Street View. We measure a Pearson correlation coefficient of -0.66 between 2012 state level carbon footprint from the transportation sector and our calculated state average MPGs. Both maps show that coastal states are greener than inland ones.}
\label{fig:mpg}
\end{figure*}
Using the output of our car classifier, we can answer diverse questions ranging from which city has the most expensive cars (New York, NY - \$11,820.62), highest percentage of Hummers (El Paso, TX - 0.39\%), highest percentage of foreign cars (San Francisco, CA - 60.02\%) or lowest percentage of foreign cars (Casper Wyoming - 21\%). We calculate these values by taking the expectation of the various car attributes across all images in each city. The average number of instances of a particular car in a region is the sum of the expected number of instances of that car across all images in the region. Thus, computing the average number of cars with a particular attribute in a city (or zip code) consists of calculating this expectation for each image in the region and aggregating this value over all images in the city (or zip code).

With $I$ an image and $c$ one of the 2,657 classes, we calculate the expected number of cars of type $c$ in a single image as
\begin{equation}
\mathbb{E}[\mathrm{\#class}\, c | I] = \sum_{\mathrm{bbox}\,b} P(\mathrm{car}|b,I) P(\mathrm{class}\, c | \mathrm{car},b,I)
\end{equation}
where bbox $b$ is the set of all detected car bounding boxes in the city (or zip code), $P(\mathrm{car}|b,I)$ is the probability of a bounding box containg a car (determined by our detection system) and $P(\mathrm{class}\, c | \mathrm{car},b,I)$ corresponds to the conditional probabilities output by the softmax layer of our CNN classifier. 

To obtain the expected value of a particular car attribute, e.g. the percentage of Hummers in a region, we aggregate category level expectations for all classes whose make is Hummer. We follow this setup in all other experiments, and note that a similar procedure can be used to find expected values for any other car attribute, including car price, miles per gallon, and country of origin.

To estimate the accuracy of our city-level car attribute predictions, we compared the distribution of cars we detect in Street View images with the distribution of registered cars in Boston, Worcester and Springfield, MA (the three Massachusetts cities in our dataset). We perform this comparison using records from Massachusetts DMV, the only state to publicly release extensive vehicle registration data~\cite{massgt}. We measured the Pearson correlation coefficient between each detected and registered make's distribution across zip codes. Twenty five of the top thirty makes have a Pearson's r correlation of r$>$0.5. Beyond Massachusetts, we measure the correlation between our detected car make distribution and the 2011 national distribution of car makes as r=0.97, verifying the efficacy of our approach. 

Car attributes can reveal aspects of a city's character that are not directly car-related. For example, our measurements show Burlington, Vermont to be the greenest city, with the highest average miles per gallon (MPG) of any city in our dataset (average MPG=25.34). Burlington is also the first city in the United States to be powered by 100\% renewable energy~\cite{burlington}. In contrast, we measured the lowest average MPG for Casper, WY (average MPG=21.28). Wyoming is the highest coal-mining state in the US, producing 39 percent of the nation's coal in 2012 and emitting the highest rate of CO\textsubscript{2} per person in the country~\cite{wyoming_coal}. We aggregate these city-level statistics by state to discover patterns across the country. For example, by mapping our detected average MPG for each state in Fig.~\ref{fig:mpg}, we can see that coastal states are greener than inland ones, a finding that agrees with published carbon footprints~\cite{state_ranking} (r=-0.66 between our calculated MPG and carbon footprints). 

\subparagraph{Income and Crime}
We can compare cities' income-based segregation levels using our car detections. After calculating the average car price for each GPS point, we measure the level of clustering between similarly priced cars using Moran's I statistic~\cite{moran} defined as
\begin{equation}
I = \frac{N \sum_{i,j} w_{i,j}(x_i - \bar{x})(x_j - \bar{x})}{\sum_{i,j}w_{i,j} \sum_{i}(x_i-\bar{x})^2}
\end{equation}
where each $x_i$ is a distinct GPS point, $\bar{x}$ is the average of $x$ and $w_{i,j}$ is a weight describing the similarity between points $i$ and $j$. We use the squared inverse Euclidean distance as a similarity metric. Moran's I of 1, -1 and 0 indicate total segregation, no segregation and a random pattern of segregation respectively. To gain further insight we visualize statistically significant clusters of expensive and cheap cars using the Getis-Ord statistic, a more local measure of spatial autocorrelation ~\cite{getis,ord1995local}. Fig.~\ref{fig:moran-i} shows our results for 22 cities with dense GPS sampling. 

Chicago is the most segregated city, with two large clusters of expensive and cheap cars on the West and East side of the city respectively. Conversely, Jacksonville is the least segregated city with a Moran’s I only 33\% as large as Chicago's and exhibits little clustering of expensive and cheap cars. As shown in Fig. ~\ref{fig:moran-i}B and Fig. ~\ref{fig:moran-i}C, Chicago's clusters of expensive and cheap cars fall in high and low income neighborhoods respectively. Our results agree with findings from the Martin Prosperity Institute~\cite{martin}, ranking Chicago, IL and Philadelphia, PA among the most segregated and Jacksonville, FL among the least segregated American cities. 

Our segregation analysis suggests that we can train a model to accurately predict a region's income level from the properties of its cars. To this end we first represent each zip code by an 88 dimensional vector comprising of car-related attributes such as the average MPG, the percentage of each body type, the average car price and the percentage of each car make in the zip code. We then use 18\% of our data to train a ridge regression model~\cite{ridge} predicting median household income from car features. Remarkably, our model achieves a city level correlation of r=0.82 and a zip code level correlation of r=0.70 with ground truth income data obtained from ACS~\cite{acs} (p$<$1e-7). 

Investigating the relationship between income and individual car attributes shows a high correlation between median household income and the average car price in a zip code (r=0.44, p$<<0.001$). As expected, wealthy people drive expensive cars. Perhaps surprisingly however, we found the most positively correlated car attribute with income to be the percentage of foreign manufactured cars (r=0.47). In agreement with our results, Experian Automotive's 2011 ranking shows that all of the top 10 car models preferred by wealthy individuals were foreign, even when the car itself was comparatively cheap (e.g. Honda Accord or Toyota Camry)~\cite{experian}. 

Following the same procedure, we predict burglary rates for cities in our test set and achieve a pearson correlation of 0.61 with ground truth data obtained from~\cite{crime}. While one of the best indicators of crime is the percentage of vans (r=0.30 for total crime against people and properties), the single best predictor of unsafe zip codes is the number of cars per image (r=0.31 and r=0.36 for crimes against people and properties respectively). According to studies conducted by law enforcement, many crimes are committed in areas with a high density of cars such as parking lots~\cite{parking}, and some departments are helping design neighborhoods with a lower number of parked cars on the street in order to reduce crime~\cite{sfcrime}. 
\section{Conclusion}
Through our analysis of 50 million images across 200 cities, we have shown that cars detected from Google Street View images contain predictive information about our neighborhoods, cities and their demographic makeup. To facilitate this work, we have collected the largest and most challenging fine-grained dataset reported to date and used it to train an ultra large scale car detection model. Using our system and a single source of visual data, we have predicted income levels, crime rates, pollution levels and gained insights into the relationship between cars and people. In contrast to our automated method which quickly determines these variables, this data is traditionally collected through costly and labor intensive surveys conducted over multiple years. And while our method uses a single source of publicly available images, socioeconomic, crime, pollution, and car related market research data are collected by disparate organizations who keep the information for private use. Our approach, coupled with the increasing proliferation of Street View and satellite imagery has the potential to enable close to real time census prediction in the future\textemdash  augmenting or supplanting survey based methods of demographic data collection in the US. Our future work will investigate predicting other demographic variables such as race, education levels and voting patterns using the same methodology.

\section{Acknowledgments}
We thank Stefano Ermon, Neal Jean, Oliver Groth, Serena Yeung, Alexandre Alahi, Kevin Tang and everyone in the Stanford Vision Lab for valuable feedback. This research is partially supported by an NSF grant (IIS-1115493), the Stanford DARE fellowship (to T.G.) and by NVIDIA (through donated GPUs).

\bibliographystyle{aaai} 
\bibliography{aaai17}
\end{document}